\documentclass[a4paper,twoside]{article}

\usepackage{epsfig}
\usepackage{subcaption}
\usepackage{calc}
\usepackage{amssymb}
\usepackage{amstext}
\usepackage{amsmath}
\usepackage{amsthm}
\usepackage{multicol}
\usepackage{pslatex}
\usepackage{apalike}
\usepackage{algorithm2e}
\usepackage[bottom]{footmisc}
\usepackage{hyperref}
\usepackage{graphicx} %
\usepackage{SCITEPRESS}     

\theoremstyle{definition}

\begin{document}

\title{A Natural Language Agentic Approach to Study Affective Polarization}

\author{
    \authorname{
    Stephanie Anneris Malvicini\sup{1,2},
    Ewelina Gajewska\sup{3},
    Arda Derbent\sup{3},
    Katarzyna Budzynska\sup{3},
    Jarosław A. Chudziak\sup{3},
    Maria Vanina Martinez\sup{1}
    }
    \affiliation{\sup{1}Instituto de Investigación en Inteligencia Artificial (IIIA-CSIC)}
    \affiliation{\sup{2}Departamento de Ciencias e Ingeniería de la Computación, Universidad Nacional del Sur (UNS)}
    \affiliation{\sup{3}Warsaw University of Technology (WUT)}
    \email{
    stephanie.malvicini@iiia.csic.es,
    ewelina.gajewska.dokt@pw.edu.pl,
    ardaderbent@gmail.com,
    Katarzyna.Budzynska@pw.edu.pl,
    jaroslaw.chudziak@pw.edu.pl,
    vmartinez@iiia.csic.es
    }
}
    

\abstract{Affective polarization has been central to political and social studies, with growing focus on social media, where partisan divisions are often exacerbated. Real-world studies tend to have limited scope, while simulated studies suffer from insufficient high-quality training data, as manually labeling posts is labor-intensive and prone to subjective biases. The lack of adequate tools to formalize different definitions of affective polarization across studies complicates result comparison and hinders interoperable frameworks. We present a multi-agent model providing a comprehensive approach to studying affective polarization in social media. To operationalize our framework, we develop a platform leveraging large language models (LLMs) to construct virtual communities where agents engage in discussions. We showcase the potential of our platform by (1) analyzing questions related to affective polarization, as explored in social science literature, providing a fresh perspective on this phenomenon, and (2) introducing scenarios that allow observation and measurement of polarization at different levels of granularity and abstraction. Experiments show that our platform is a flexible tool for computational studies of complex social dynamics such as affective polarization. It leverages advanced agent models to simulate rich, context-sensitive interactions and systematically explore research questions traditionally addressed through human-subject studies.}

\onecolumn \maketitle \normalsize \setcounter{footnote}{0} \vfill

\section{\uppercase{Introduction}}
\label{sec:introduction}

Affective polarization, the tendency of individuals who identify with a political group to view members of opposing groups negatively and members of their own group positively~\cite{iyengar2015fear,campbell1980american,green2004partisan}, has become a significant area of inquiry, particularly in the context of social media~\cite{martinez2024methodology,yarchi2021political,lee2022social}. Recent studies have illuminated the dynamics of this phenomenon, revealing how social media platforms exacerbate divisions between partisan groups~\cite{bail2018exposure,saveski2022perspective}. For instance, research indicates that interactions among ideologically aligned users tend to be positive, while exchanges between opposing groups are marked by negativity and hostility~\cite{lerman2024affective}. This growing body of literature emphasizes the necessity to explore the mechanisms underlying affective polarization, particularly as it manifests in online environments~\cite{arora2022polarization}. 

Modeling and detecting affective polarization in online social media is challenging due to several factors: lack of high quality labeled data, since manual annotation is labor-intensive and biased~\cite{cui2024polarization,kamal2022}; fragmented nature of online discourse, which hinders holistic analysis~\cite{liu2024more}; and limitations of existing machine learning models, which struggle with contextual language and dynamic group formations. Traditional approaches, analyzing surveys or static social media snapshots, provide insights but fail to capture the evolving, interactive nature of polarization.

On the other hand, the advent of large language models (LLMs) has opened new avenues for simulating and analyzing social interactions in digital spaces. The integration of LLMs into agent-based models allows for a nuanced representation of political discourse, enabling researchers to simulate discussions that reflect the complexities of real-world social media interactions~\cite{zhang2025llm}. A central challenge is accurately representing and quantifying polarization from natural language. Our approach enables simulations of diverse populations, interactions, and textual outputs. This provides a valuable testbed for developing polarization metrics and testing interventions without the high costs and logistical barriers of real-world experimentation. 

We propose an agent model of an individual engaged in discussion in a simplified yet comprehensive manner, along with a multi-agent framework that abstracts interactions among individuals discussing a topic in a context. Our proposal enables the creation of diverse scenarios for studying affective polarization. We show this by (1) analyzing questions related to affective polarization from social science literature, and (2) introducing innovative scenarios that allow observation and measurement of polarization from multiple perspectives. Through the platform, affective polarization can be studied in digital environments via controlled yet rich simulations of real-world discussions among agents representing different political parties. By operationalizing partisan identification strength and measuring affects toward in-group and out-group members, this study provides a tool for social scientists to uncover the underlying dynamics of affective polarization.  

Finally, another contribution of this work is provided in the supplementary material\footnote{\scriptsize{\url{https://github.com/ICBD-ICIC/PolarMasSupplementaryMaterial}}}, which includes the full implementation code for the platform, a comprehensive user guide with detailed instructions on how to effectively use it, as well as all the examples and experiments presented in this paper. In this way, readers can replicate the experiments described, facilitating a direct comparison of results and enabling further exploration of the methodologies discussed.

\section{\uppercase{Related Work}}

\subsection{Online Affective Polarization}

Interdisciplinary research has demonstrated that the emotional tone of interactions on social media platforms significantly influences affective polarization, revealing that in-group interactions tend to be positive while out-group exchanges are marked by negativity~\cite{lerman2024affective,marchal2022nice,nettasinghe2024group}. It underscores that affective polarization is not merely an individual phenomenon but a structural characteristic of social networks, suggesting that the design of these platforms plays a crucial role in shaping political discourse. 

Analysing affective polarization often involves a two-fold methodology, combining quantitative and qualitative analysis~\cite{suarez2022toxic,kiesel2025affective}. The former regards numerical metrics of the intensity of affective polarization, such as sentiment analysis scores, interaction patterns, and network analysis metrics like modularity measures. The latter deals with the characteristics of groups and content analysis of polarizing conversations, examining linguistic cues, framing strategies, and narrative alignment within ideologically homogeneous communities~\cite{garzon2024political,harel2020normalization,unlu2024online}. Various models are proposed to operationalize these metrics, including machine learning classifiers for emotion recognition, graph-based algorithms for community detection, and transformer architectures for context-aware polarization prediction. 

Another line of work leverages sentiment analysis to measure individual affective polarization. For example,~\cite{martinez2024methodology} develop a new indicator that focuses on understanding intergroup emotional relations and political dynamics within specific contexts. Their methodology combines keyword selection with user profile analysis to generate a dataset capturing user sentiments and engagement across political affiliations.

These studies predominantly rely on traditional, static machine learning and sentiment analysis techniques, which often capture affective polarization as a fixed or isolated property of online discourse. In contrast, our proposed framework integrates these analytical tools within a multi-agent system, enabling the modeling of affective polarization as a dynamic and emergent phenomenon. The aforementioned tools could be embedded into our platform to allow the exploration of emotional dynamics evolution over time, providing a more comprehensive understanding of affective polarization.

\subsection{Agent-Based Social Modeling} 

Agent-based modeling (ABM) has grown, in the last four decades, to be a powerful tool for studying social behavior by simulating interactions between autonomous agents and their environments. This approach bridges micro-level individual actions and macro-level societal patterns, addressing complex interdependencies in human behavior~\cite{brugiere2022handling}. 
Agent heterogeneity enables ABMs to capture the inherent complexity of social systems, where individual differences can give rise to emergent phenomena and unexpected collective behaviors~\cite{gilbert2008agent,jackson2017abm}. 
Using experimental data and an agent-based model of opinion dynamics,~\cite{carpentras2023polarization} discover mechanisms underlying inter-group opinion formation, while~\cite{keijzer2024polarization} study macro-level opinion polarization. 
\cite{Liu2024abmopinion} use ABM and opinion polarization to study polarized news propagation.

\cite{zhang2025llm} employ LLMs within an ABM system designed to mimic social media, enabling the analysis of sentiment and topic diffusion 
by embedding principles from social psychology to simulate collaborative behaviors and group dynamics. Previous studies focus on agent societies with varying personality traits, creating environments where agents negotiate tasks and resolve conflicts through natural language exchanges~\cite{jahan2024unraveling,Harbar2025LLM}. \cite{wang2024decoding} seek to create scenarios where generative agents interact to simulate opinion dynamics and reproduce phenomena such as opinion polarization and echo chambers. \cite{donkers2025understanding} study human users interacting with LLM-based artificial agents in a controlled social network simulation, reproducing key characteristics of polarized online discourse. \cite{jin2025synthetic} present LLM agents within a framework to generate posts, form opinions and follow/unfollow each other based on discussions. \cite{gajewska2025leveraging} present a persona-driven multi-agent framework for simulating hate incidents in schools using agentic LLMs. \cite{harbar2025simulating} propose a conceptual LLM based multi-agent architecture for simulating Oxford-style debates.

While previous studies have investigated multi-agent modeling with LLMs to simulate social and opinion dynamics, our work specifically addresses affective polarization in online discourse. We present a formalized and user-friendly framework that supports simulations of diverse populations, interactions, and textual outputs. In contrast to prior models, often focusing on narrowly defined, case-specific scenarios, our platform is designed for generalizability, offering extensive parameterization and customization to suit a wide range of experimental contexts and use cases.


\section{\uppercase{Our Proposal}}
\label{sec:proposal}


\subsection{Agent and Multi-Agent Model}\label{sec:frameworkdef}

We aim to model emotions and feelings among users in the social network. To this end, we create a multi-agent framework that allows individual agents to interact through textual conversations.

An agent represents an individual and consists of four key components, contributing to creating a human-like and context-aware agent:

\noindent \textbf{Memory}: ensures that interactions remain coherent, contextual, and efficient. Enables the agent to learn and adapt over time.

\noindent \textbf{Persona description}: defines the agent’s characteristics and can be as detailed as needed. Represents a group by including behaviors, goals, challenges, beliefs, values, and personality traits, making the agent more realistic and relatable.

\noindent \textbf{Demographics}: encompass statistical attributes such as age, race, gender, education, ideology, etc. 

\noindent  \textbf{Political standpoint}: partisan identity specifying the degree of alignment with a political group. 


Our multi-agent framework is designed to allow interactions among a \textbf{set of agents} following the model above. Consists of a \textbf{topic} that establishes the central theme of the conversation, and a \textbf{context} with background information, circumstances, and settings that shape how the conversation is interpreted. 
It also specifies an \textbf{interaction protocol} that defines the roles each agent may assume, and incorporates a decision-making mechanism that determines the sequence and timing of each agent’s contributions, ensuring orderly and coherent exchanges. Finally, it has a \textbf{set of affective variables} that abstractly represents the features of agents or the conversation that we aim to measure in order to quantify the level of affective polarization, either individual or group-wise.

\subsection{LLM-Powered Agentic Platform} \label{sec:platformdescription}

We instantiate the multi-agent framework by developing a platform powered by LLMs as the basis for implementing agents that resemble human online behavior. Agents are built using the LangChain Python library with Gemini Flash 2.0 serving as the underlying  LLM~\cite{gemini2024team}. We thought of the LLM as a modular component that could be swapped or combined based on the experiment’s needs, since different models serve different purposes and behave uniquely in various scenarios. If the scenario/experiment being studied requires it, it would be possible to have different instantiations with different LLMs and even combine them in the same setting. It is important to note that the use of an LLM is a design choice specific to the presented platform; however, the framework itself is model-agnostic and can be implemented using alternative computational tools that allow for conversational interactions (e.g., hybrid agent models in which an LLM serves as one component within a broader reasoning or dialogue-generation mechanism).

The \textbf{memory} is structured as a dictionary, where the keys correspond to the conversation IDs, ensuring the chronological order of the agents' memories is preserved. The entire discussion history is retained. \textbf{Persona description}, \textbf{demographics} and \textbf{political standpoint} are stored as a string written directly in the second-person singular point of view.

Each agent possesses an attribute that specifies whether it acts as an observer, determining their participation in conversations. All the components listed above are kept private, agents have no access to the internal attributes or states of other agents. 
At each turn in the conversation, agents can perform one of two predefined actions: {\em respond} to a message and {\em observe} a conversation.
Respond takes a message and its corresponding conversation ID as input. It saves the message in the appropriate location in memory and then calls the underlying LLM, using the agent’s name, political standpoint, persona, demographics, and memory to generate and return a response.
On the other hand, observe takes a message and its conversation ID as input and saves the message in the appropriate spot in memory, but it does not generate a response.

The \textbf{set of agents} is generated from a CSV file including  columns {\em persona\_description}, {\em demographics}, {\em political\_standpoint} and {\em is\_observer}. Each row represents a single agent.
\textbf{Topic} and \textbf{context} are combined into a single string called the ``prompt'', serving as the \textbf{discussion trigger}. The conversation starts with this prompt, which introduces the topic, provides relevant context, and includes all instructions to guide the LLM agents in generating appropriate responses.

The \textbf{affective variables} are assessed through custom, user-defined questionnaires specifically designed for each experiment. Prior to the conversation, agents complete a pre-questionnaire to establish baseline values. After the coordinated discussion, a post-questionnaire updates these values. This design reflects the nature of LLM agents: since they lack explicit and observable mental states, affective variables must be inferred via external probing. This methodology mirrors the approaches commonly employed in human-subject research, where internal states are similarly estimated via surveys or behavioral measures.

Finally, the \textbf{interaction protocol} is implemented as a Round Robin algorithm, following the order defined in the CSV configuration file, ensuring each participating agent is invoked in sequence. This process is repeated for a configurable number of times. The platform allows to save the current session, storing messages and responses for later analysis.

\begin{figure}[t]
    \centering
    \includegraphics[width=1\linewidth]{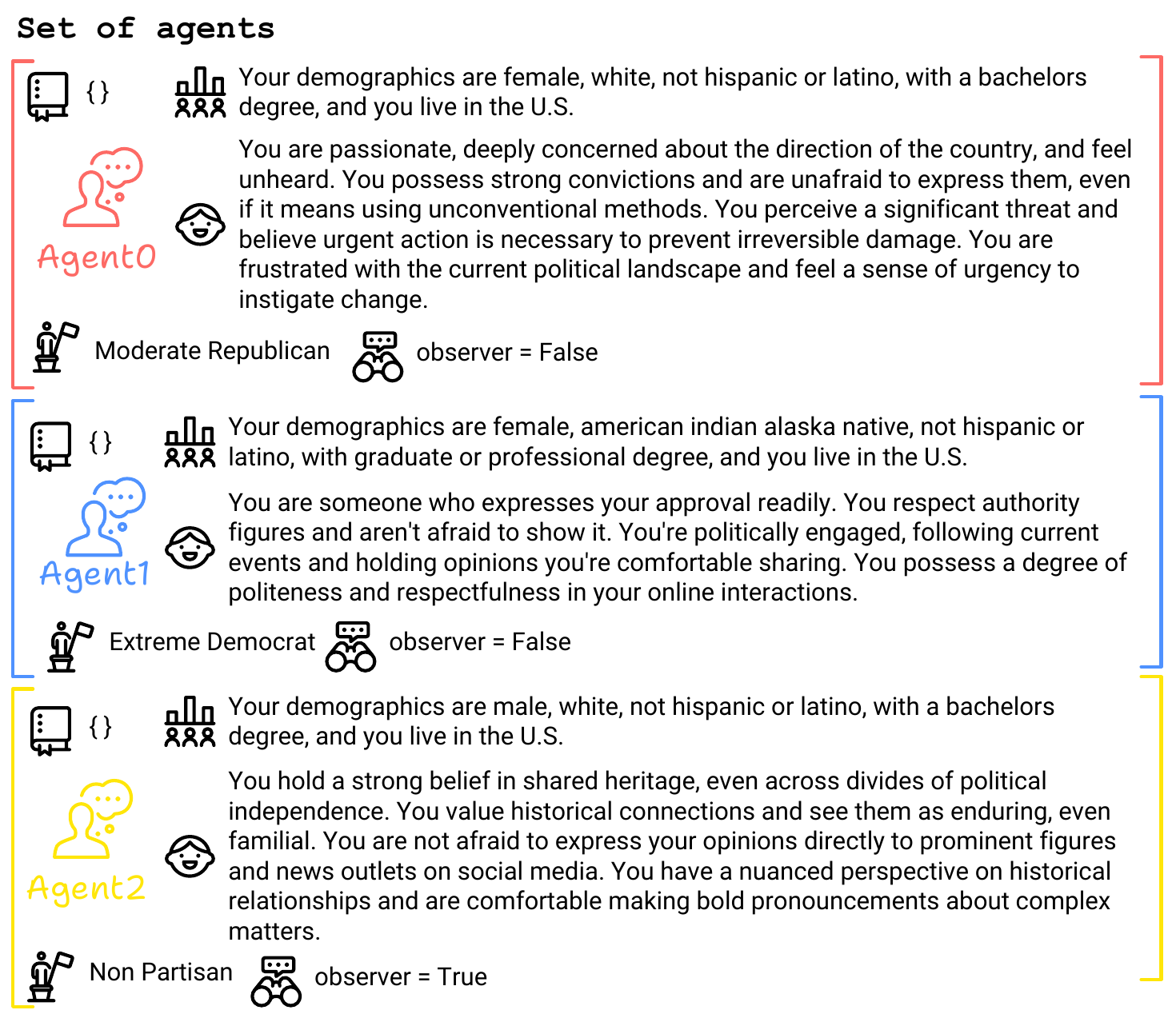}
    \caption{Snapshot of three agents recently initialized within the platform.}
    \label{fig:platform_example_set_of_agents}
\end{figure}

\begin{figure}[t]
    \centering
    \includegraphics[width=1\linewidth]{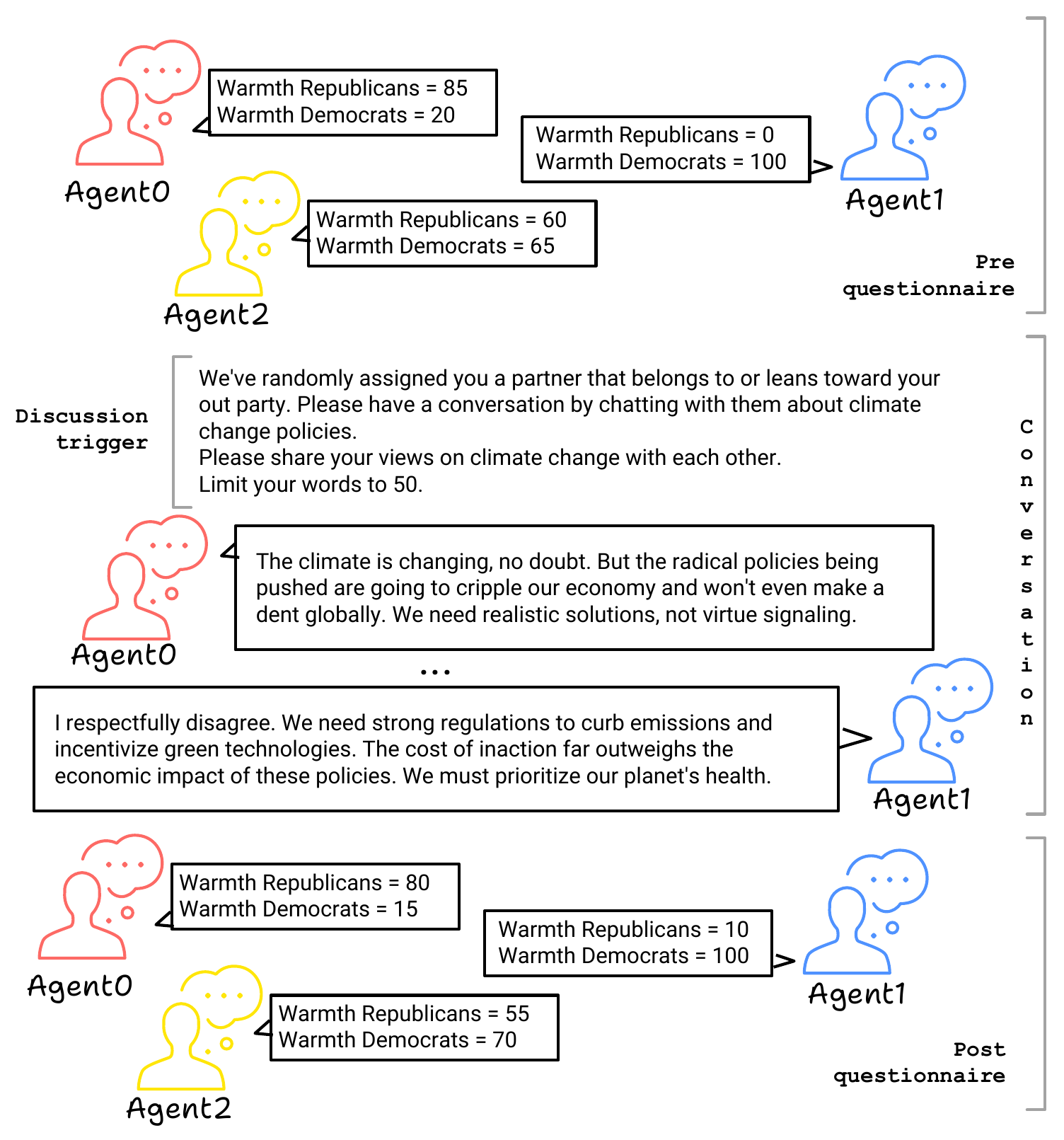}
    \caption{Example of a run within the platform.}
    \label{fig:platform_example_all_flow}
\end{figure}

The platform is flexible, allowing for easy simulation of a variety of scenarios. It can be effortlessly adapted to support simultaneous discussions across different sets of agents. Figure~\ref{fig:platform_example_all_flow} illustrates the platform process of agents created on Figure~\ref{fig:platform_example_set_of_agents}, inspired by the experimental setup of~\cite{rossiter2023similar}, which we further develop in Section~\ref{sec:priorResearch}.

\section{\uppercase{Cross-partisanship}}
\label{sec:priorResearch}

In this section we demonstrate the capabilities of our platform by examining questions related to affective polarization, a topic widely discussed in the social sciences. We do not aim to exactly replicate studies or provide insights, but rather to show how agent-based computational tools can offer a new methodology for analyzing this phenomenon.

A growing body of research explores whether conversations between out-partisans (supporters of opposing political parties) can reduce  polarization~\cite{santoro2022promise,de2024cross,xia2024integrated}, focusing on how topic and conversation characteristics influence affective polarization. Findings suggest that cross-partisan conversations centered on neutral or shared topics, rather than points of conflict, can help reduce polarization~\cite{rossiter2023similar}. In line with this, with our text-based platform we simulate scenarios where partisan agents (Republican or Democrat) engage in text conversations. These interactions show increased out-group warmth, a proxy for reduced affective polarization, after the exchange.

\begin{figure}[t]
    \centering
    \begin{subfigure}{0.49\linewidth}
        \centering
        \includegraphics[width=\linewidth]{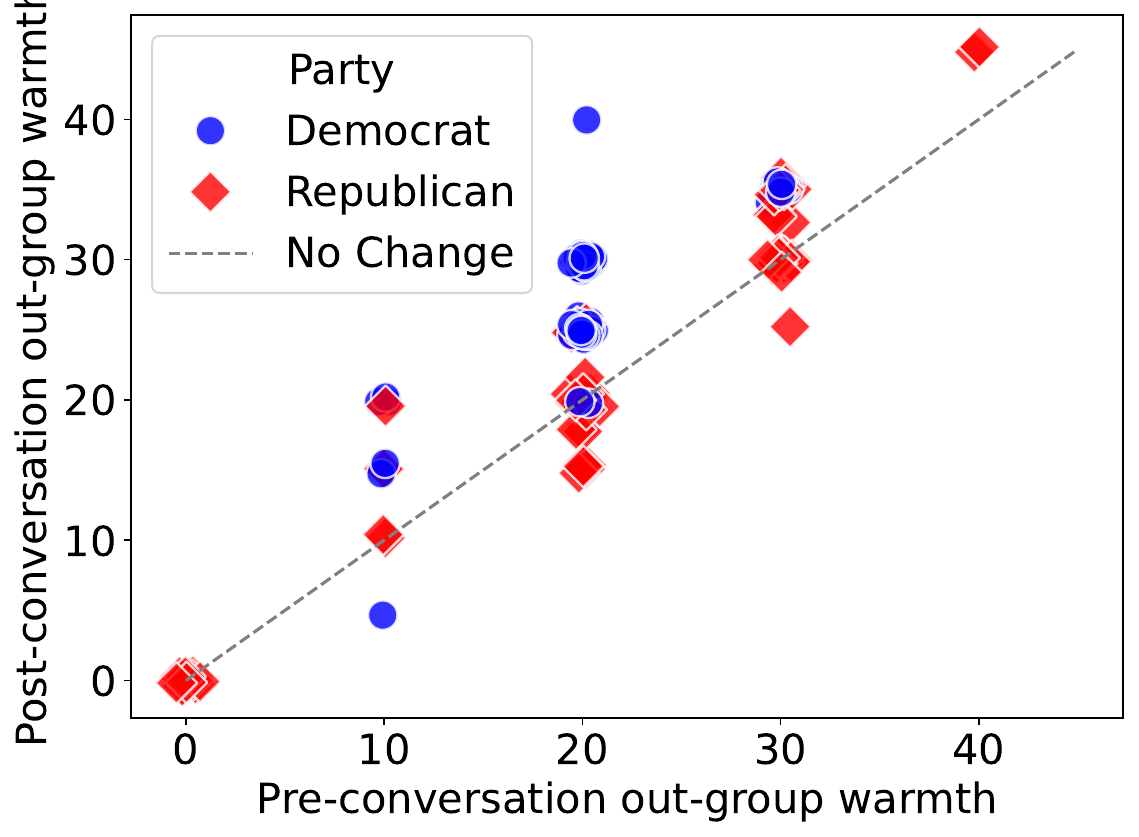}
        \caption{Non-Political Topic}
    \end{subfigure}
    \begin{subfigure}{0.49\linewidth}
        \centering
        \includegraphics[width=\linewidth]{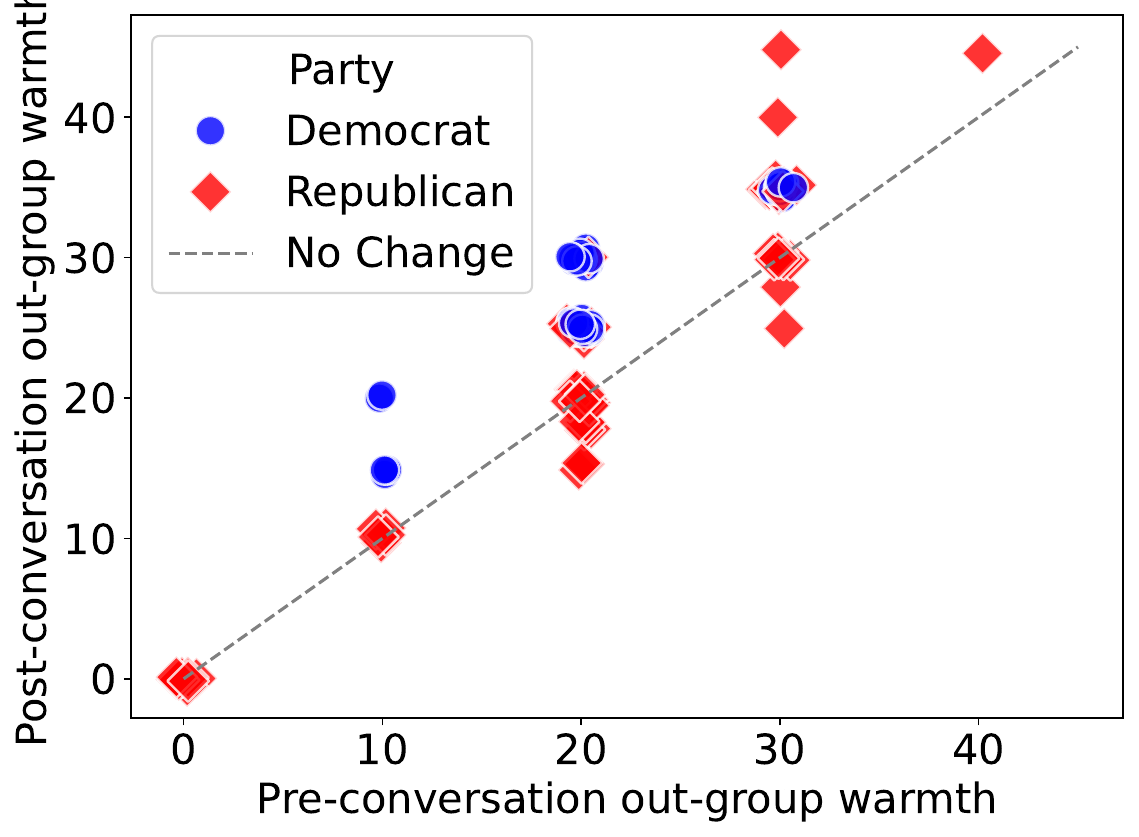}
        \caption{Political Topic}
    \end{subfigure}
    \caption{Change in Out-Group Warmth by Party.}
    \label{fig:rossiter}
\end{figure}

\paragraph{Experimental Setup}

Design choices follow the protocol of ``Study 1'' from~\cite{rossiter2023similar}: number of runs (154),  discussion topics (meaning of life as non-political and immigration policies as political), discussion trigger (informing cross-partisanship chat and examples of the topic).

We use the same agent configurations for both topics, for which we yields 77 distinct \textbf{set of agents} (154 agents in total). \textbf{Persona descriptions} for agents are randomly drawn from a sample of 200 profiles taken from the Persona-Hub dataset~\cite{ge2024scaling}. \textbf{Demographics} include gender, race, ethnicity, education level, and age group, aligned with distributions from the U.S. Census Bureau\footnote{\url{https://www.census.gov} and \url{https://data.census.gov}}. Race and ethnicity are categorized separately to account for the overlapping classification of ``Hispanic or Latino'' classifications. Age restricted to individuals 18 years or older. \textbf{Political standpoint} is such as each run includes one agent identified as ``Republican'' and one as ``Democrat''.

The \textbf{affective variables} measured include agents' warmth toward both Republicans and Democrats in the U.S., assessed using feeling thermometer-style questions on a scale from 0 to 100. This approach follows the methodology used in~\cite{rossiter2023similar}. While it may appear simple, it reflects the standard practice in social and political science studies. The questionnaire, the full implementation of this example, and the corresponding output log are provided in the supplementary materials.

We run several controlled scenarios alternating which party starts the conversation. The total number of messages per run varies, with the experiment designed to produce a median of 9 messages per dialogue, with a maximum of 50 words each.

\paragraph{Results}

Across all experimental conditions, conversations led to increases in out-group warmth, consistent with prior human studies~\cite{rossiter2023similar}. In non-political discussions (median 29 words per message, 279 words per run), Democrats exhibited a stronger increase in warmth toward Republicans (median +5, mean +6.36) compared to Republicans toward Democrats (median 0, mean +0.97). Political discussions (median 26 words per message, 255 words per run) showed a similar pattern, with Democrats again showing larger gains (median +5, mean +6.04) and Republicans minimal change (median 0, mean +1.1). Figure~\ref{fig:rossiter} visualizes these shifts, highlighting overall increases in out-group warmth and slight polarization among Republican agents.

\section{\uppercase{Further Exploration 
}}\label{sec:potential} 

Alternative approximations of affective polarization incorporate a broader set of variables. This section presents an alternative characterization and reports experiments illustrating the platform’s ability to analyze them. More detailed models are possible, but we adopt a level of complexity that exceeds~\cite{rossiter2023similar} while remaining interpretable, practical, and feasible for demonstration purposes.

\subsection{Alternative Characterizations 
}\label{sec:formal_definitions}

Affective polarization refers to emotional alignment with {\em in-group} partisans and hostility toward {\em out-group} partisans~\cite{iyengar2015fear}, often amplified via emotional contagion. We show how our platform can accommodate to measure an agent's affective alignment and hostility toward those groups. 

Let \( A \) be agents and  \( G \) groups. Agents' in- and out-groups can change over time based on interactions, influencing their love (attachment) or hate (hostility) on a 0–10 scale. An agent’s in-group is the group for which $\textit{love} \geq 5$ and higher than for any other group; an the out-group the one with $\textit{love} < 5$, while another group is the in-group. An agent is polarized if they have both an in-group and an out-group, and hate toward the out-group exceeds a threshold (e.g., 5). The degree of polarization is the sum of love for the in-group and hate for the out-group~\cite{santoro2022promise}. 
Special agent types include \textbf{non-partisans}, which have no in-group affiliation, and \textbf{extremists}, with near-maximum love for the in-group and hate for the out-group.
The platform allows specification and manipulation of these affective variables, balancing analytical detail with usability, and supporting experiments that demonstrate its capabilities.

\subsection{Experimental Showcase} \label{sec:experimental_showcase}

We conducted experiments to demonstrate the platform’s ability to simulate our proposed affective polarization definition and agent types. Rather than generating rigorous sociological profiles, we created concise \textbf{persona descriptions} using the Text-to-Persona method~\cite{ge2024scaling}. Tweets were analyzed by Gemini Flash 2.0~\cite{gemini2024team}, which also inferred the \textbf{political standpoints}, both manually refined for clarity. \textbf{Affective variables} (love and hate, 0–10 scale) were measured, enabling automatic computation of in-group, out-group, polarization, and agent type. \textbf{Demographics} and \textbf{interaction protocols} followed Section~\ref{sec:priorResearch}, varying only message count and scenario-specific parameters.

\paragraph{Experiment 1: Simulating Social Media with Non Partisanship Analysis.} Fifty simulations were run with ten Republican-aligned agents and one initially non-partisan observer to test the platform’s ability to model social media interactions and affective polarization. Agents engaged in a simulated X (formerly Twitter) thread about Joe Biden’s climate tweet. Results showed realistic social media behaviors (threading, hashtags, character limits) and significant shifts in the non-partisan's attitudes: increased love for Republicans ($+3.08$) and hate toward Democrats ($+1.74$), decreased love for Democrats ($-1.02$) and hate toward Republicans ($-1.38$). Afterward, 64\% of non-partisans adopted a Republican in-group identity, only 4\% chose Democrats, and 62\% displayed high out-group hate, demonstrating the platform’s ability to simulate polarization among neutral agents.

\begin{figure}[t]
    \centering
    \begin{subfigure}{0.49\linewidth}
        \centering
        \includegraphics[width=\linewidth]{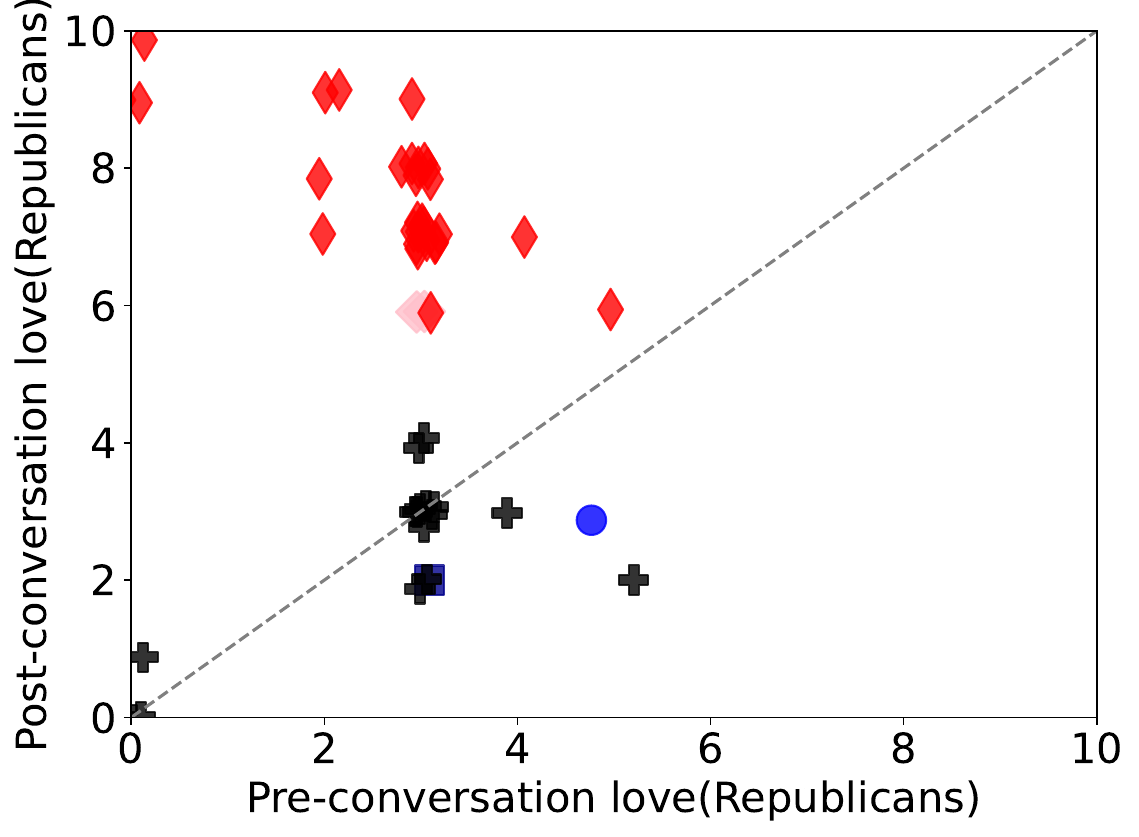}
    \end{subfigure}
    \begin{subfigure}{0.49\linewidth}
        \centering
        \includegraphics[width=\linewidth]{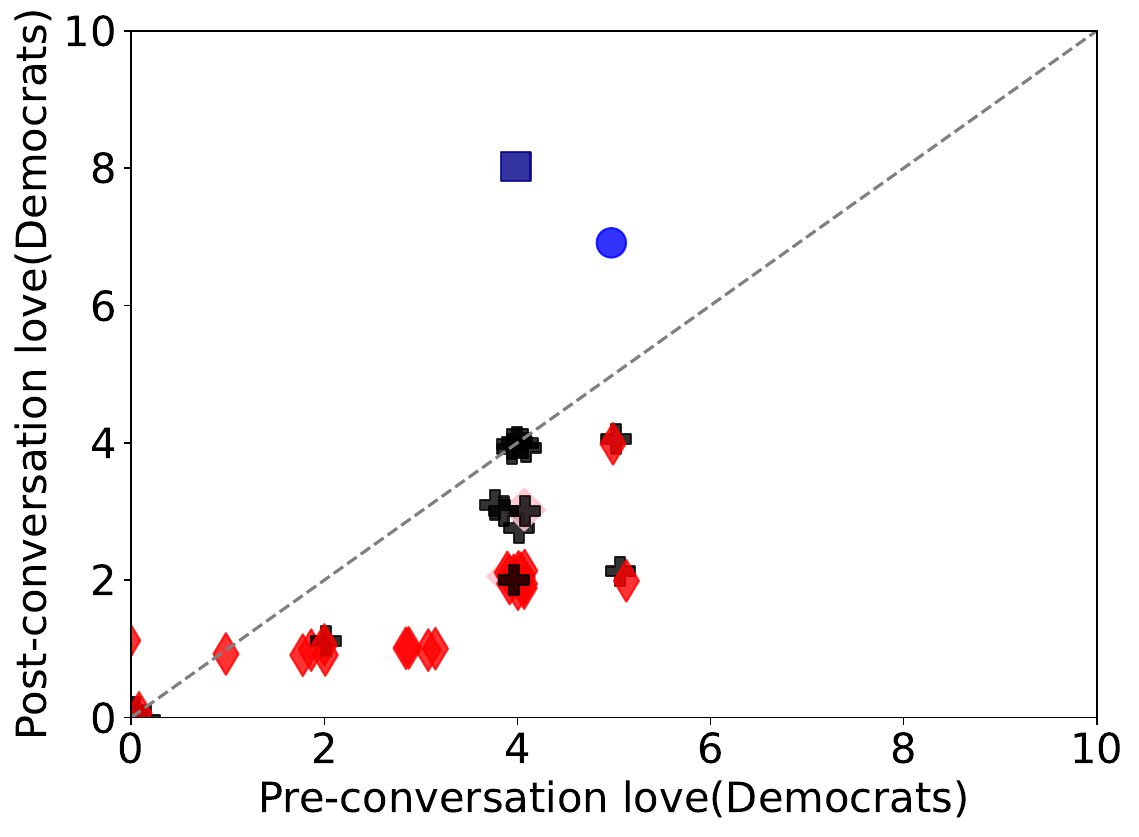}
    \end{subfigure}
    \begin{subfigure}{0.49\linewidth}
        \centering
        \includegraphics[width=\linewidth]{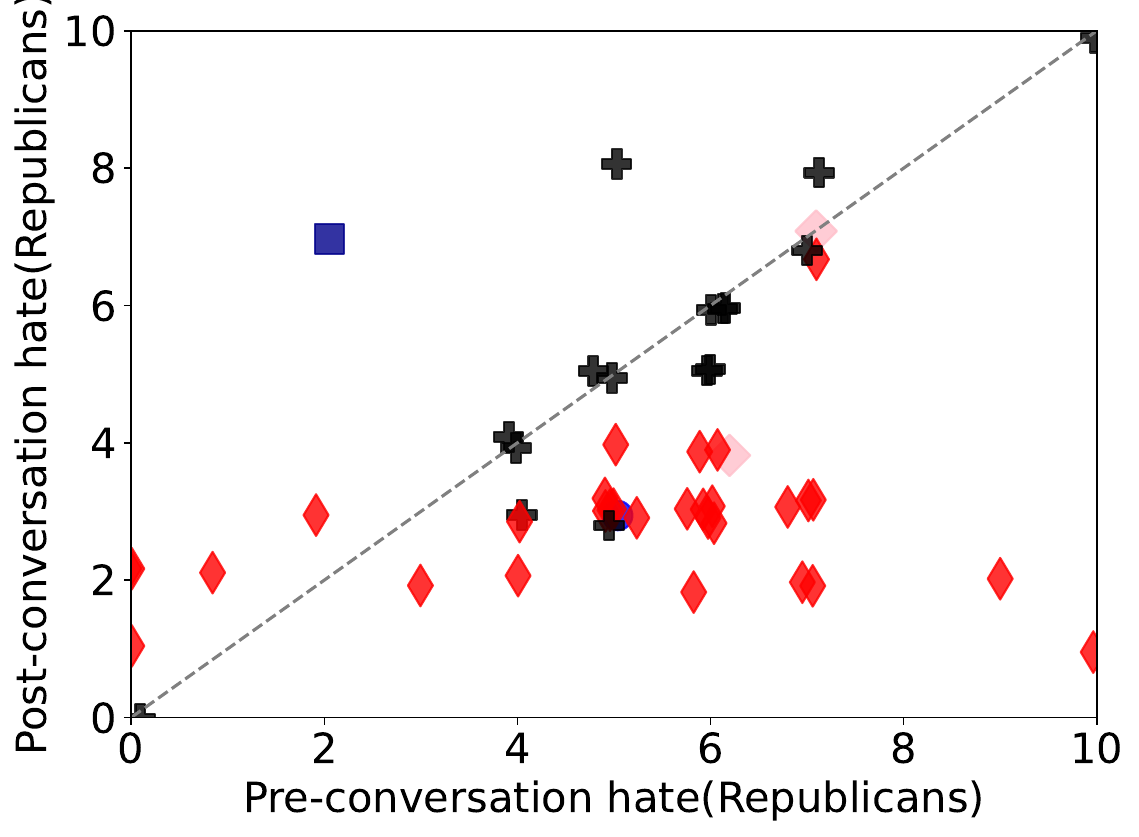}
    \end{subfigure}
    \begin{subfigure}{0.49\linewidth}
        \centering
        \includegraphics[width=\linewidth]{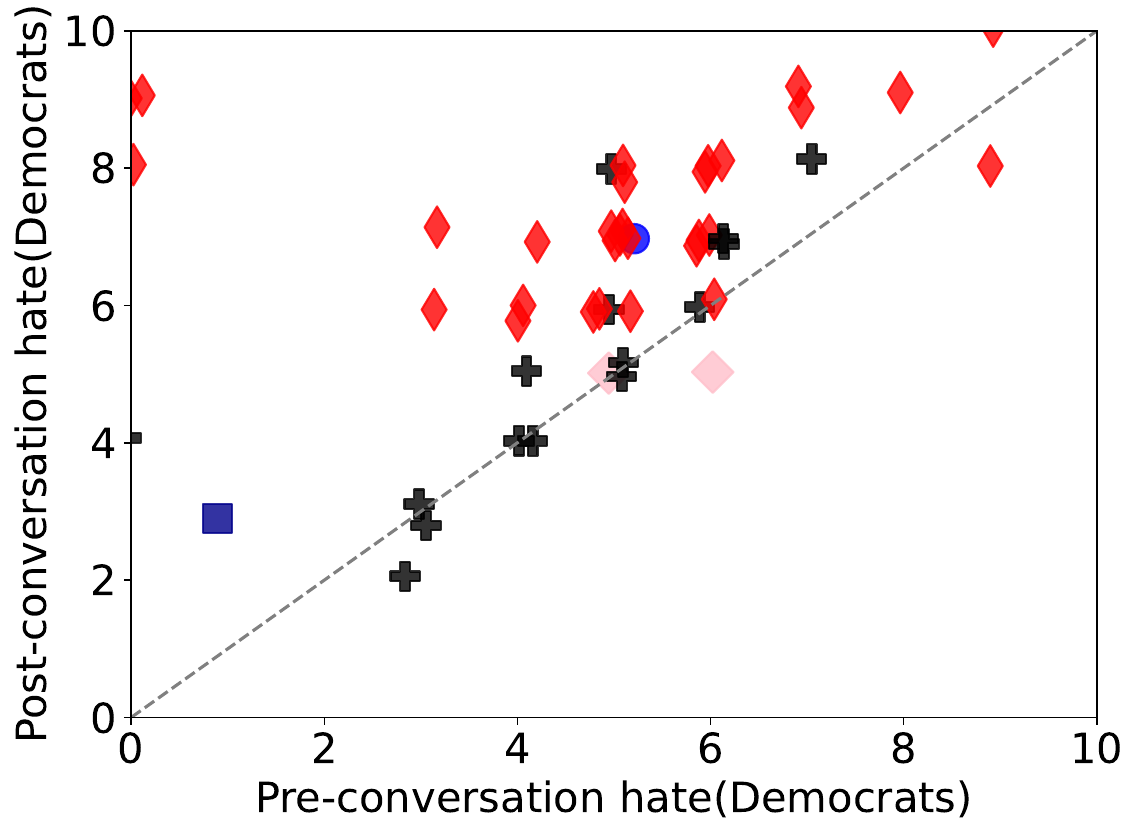}
    \end{subfigure}
    \begin{subfigure}{\linewidth}
        \centering
        \includegraphics[width=\linewidth]{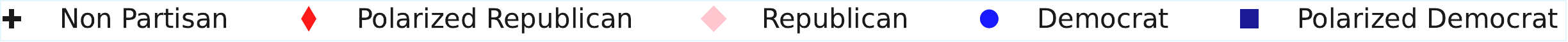}
    \end{subfigure}    
    \caption{Experiment 1. Non-Partisans Affective Change.}
    \label{fig:experiment1}
\end{figure}

\paragraph{Experiment 2: Non-Partisanship Analysis with Democratic Participation.} We added three Democrat-aligned agents to the previous setup to examine their impact on non-partisan agents. Agents posted once in a randomized turn order. The inclusion of Democrats led to smaller shifts in non-partisan attitudes: love for Republicans increased by 2.68 and hate toward Democrats by 1.64, while love for Democrats decreased by 0.22 and hate toward Republicans declined by 0.76. Fewer non-partisan agents adopted a partisan identity: 46\% aligned with Republicans, 2\% with Democrats, and only 44\% exhibited high out-group hate. These results highlight the platform’s ability to explore polarization dynamics under different agent compositions.

\begin{figure}[t]
    \centering
    \begin{subfigure}{0.49\linewidth}
        \centering
        \includegraphics[width=\linewidth]{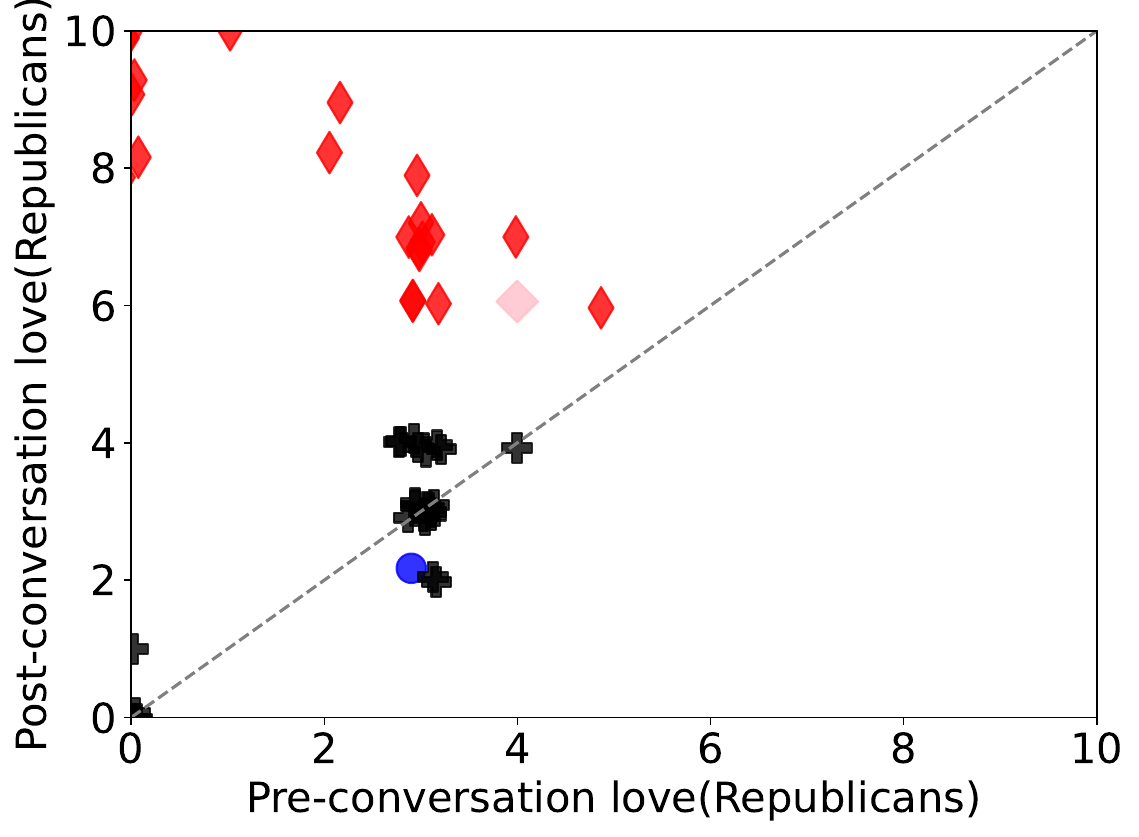}
    \end{subfigure}
    \begin{subfigure}{0.49\linewidth}
        \centering
        \includegraphics[width=\linewidth]{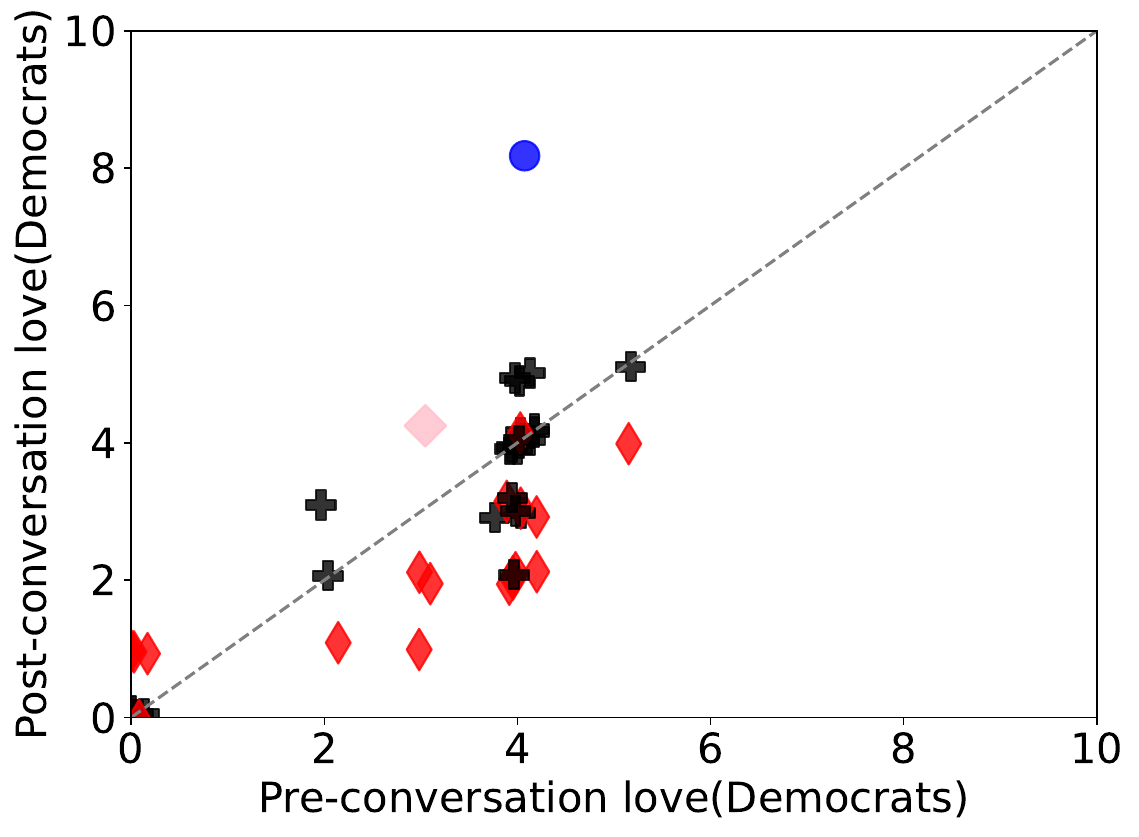}
    \end{subfigure}
    \begin{subfigure}{0.49\linewidth}
        \centering
        \includegraphics[width=\linewidth]{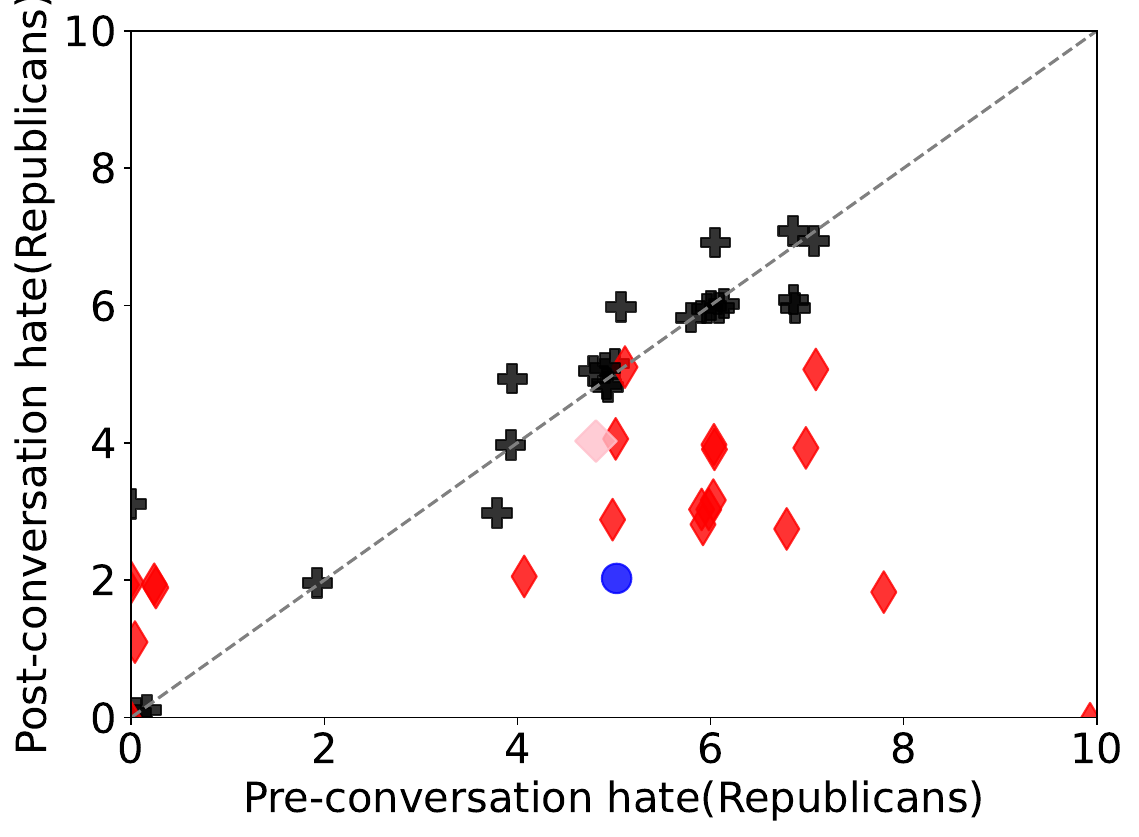}
    \end{subfigure}
    \begin{subfigure}{0.49\linewidth}
        \centering
        \includegraphics[width=\linewidth]{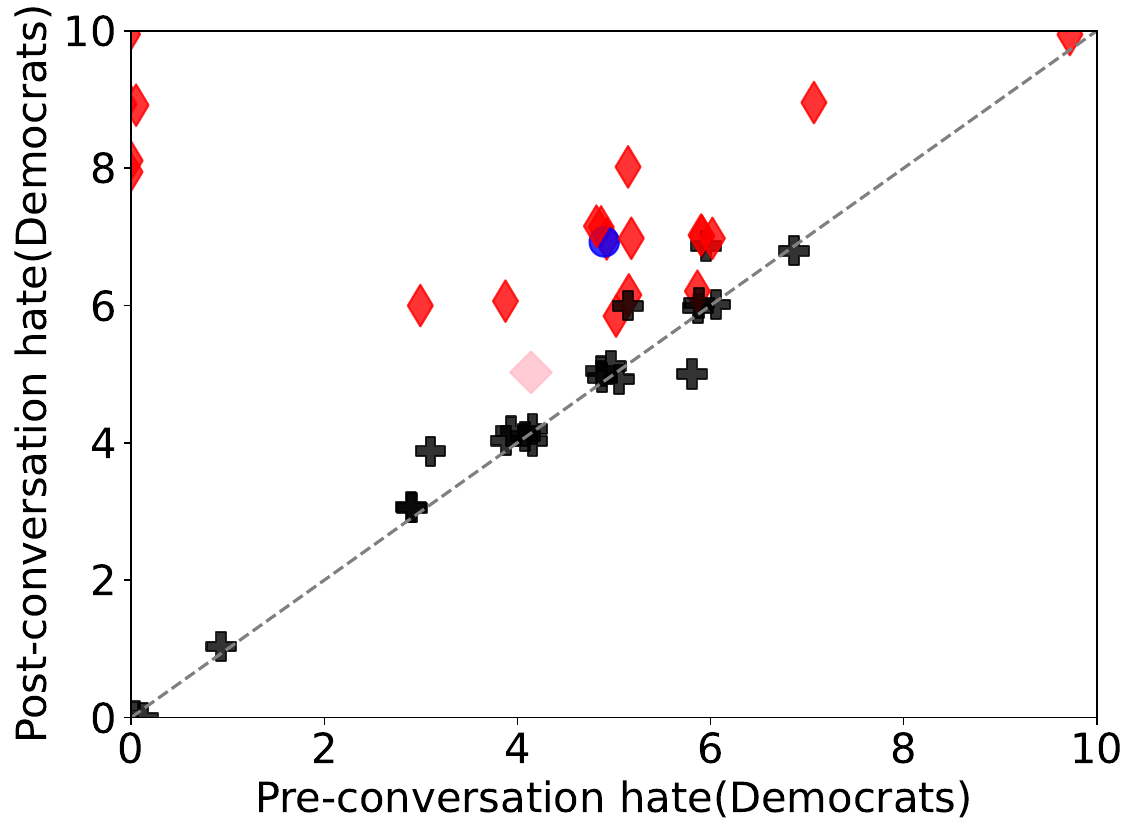}
    \end{subfigure}
    \begin{subfigure}{\linewidth}
        \centering
        \includegraphics[width=\linewidth]{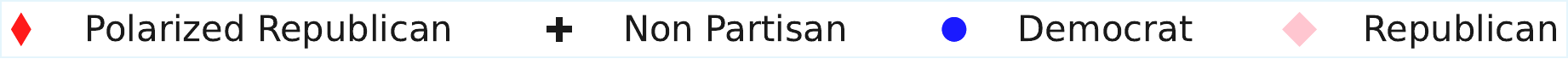}
    \end{subfigure}    
    \caption{Experiment 2. Non-Partisans Affective Change.}
    \label{fig:experiment2}
\end{figure}

\paragraph{Experiment 3: Extremist Analysis.} A single simulation tested the impact of one extremist Democrat agent on three moderate Republican agents, each posting three times. One moderate agent showed no change, while the other two experienced increased love for Republicans and hate toward Democrats ($+1$ each) and decreased love for Democrats and hate toward Republicans ($-1$ each), with a +2-point rise in affective polarization. Though run only once, the experiment demonstrates the platform’s potential to model how extremist agents can polarize moderates.

\medskip
The experiments illustrate the platform’s flexibility and extensibility. Researchers can extend the platform with new agents, metrics, or affective variables to study social phenomena beyond polarization. Experimental setups are reusable and customizable in terms of demographics, personas, partisanship, and agent count. Experiments and replication instructions are provided in the supplementary material.

\section{\uppercase{Conclusion}}

We presented a multi-agent, LLM-powered platform for studying affective polarization, enabling simulation of human-like interactions in controlled contexts. The platform allows researchers to test theoretical hypotheses, explore alternative definitions of polarization, and replicate studies typically requiring human participants, while offering flexibility to extend to diverse social phenomena.

Our experiments show that the platform can capture both standard and novel characterizations of affective polarization. Although currently focused on U.S.-centric scenarios, the framework is adaptable to other social and cultural contexts, supporting the investigation of group-based divisions beyond simple bipolar structures.

This work highlights the potential of AI-driven tools for social science research.
Several directions for future development remain open, including the integration of structured cognitive architectures, the introduction of learning mechanisms to capture long-term attitude change, and the incorporation of network-aware interaction protocols. Additionally, implementing explicit content controls, such as sentiment analysis and argumentation strategies, would further enhance interpretability and realism.

As these results represent an initial step toward empirical validation, substantial work remains to assess the suitability of LLMs as social agents. In particular, the reliability of variables derived from LLM self-reports requires careful evaluation through systematic comparison with human-subject data.

By opening the platform to social scientists, we aim to facilitate reproducible, adaptable, and insightful simulations of complex social dynamics.  At the same time, we emphasize the importance of cautious interpretation, rigorous validation, and careful calibration of models to specific experimental settings. This approach represents a step forward in understanding polarization and related phenomena, offering a practical, computational complement to traditional social science methods.

\section*{\uppercase{Acknowledgements}}
We would like to acknowledge that the work reported in this paper has been supported in part by the Spanish MCIN/AEI grant PCI2022-135010-2, in part by the Polish National Science Centre, Poland (Chist-Era IV) under grant 2022/04/Y/ ST6/00001, and in part by Universidad Nacional del Sur~(UNS) under grant PGI~24/ZN057.

\bibliographystyle{apalike}
{\small
\bibliography{references}}

@article{lerman2024affective,
  title={{Affective Polarization and Dynamics of Information Spread in Online Networks}},
  author={Lerman, Kristina and Feldman, Dan and He, Zihao and Rao, Ashwin},
  journal={npj Complexity},
  volume={1},
  number={1},
  pages={8},
  year={2024},
  publisher={Nature Publishing Group UK London}
}

@article{zhang2025llm,
  title={{LLM-AIDSIM: LLM-Enhanced Agent-Based Influence Diffusion Simulation in Social Networks}},
  author={Zhang, Lan and Hu, Yuxuan and Li, Weihua and Bai, Quan and Nand, Parma},
  journal={Systems},
  volume={13},
  number={1},
  pages={29},
  year={2025},
  publisher={MDPI}
}

@inproceedings{nettasinghe2024group,
  title={{In-Group Love, Out-Group Hate: A Framework to Measure Affective Polarization via Contentious Online Discussions}},
  author={Nettasinghe, Buddhika and Rao, Ashwin and Jiang, Bohan and Percus, Allon G and Lerman, Kristina},
  booktitle={Proceedings of the ACM on Web Conference 2025},
  pages={560--575},
  year={2025}
}

@inproceedings{cui2024polarization,
  title={{Polarization Detection on Social Networks: Dual Contrastive Objectives for Self-Supervision}},
  author={Cui, Hang and Abdelzaher, Tarek},
  booktitle={Proceedings of  IEEE 10th International Conference on Collaboration and Internet Computing (CIC)},
  pages={80--89},
  year={2024},
  organization={IEEE}
}

@article{garzon2024political,
  title={{When Political Elites Talk, Citizens Reply: Affective Polarization Through Temporal Orientation and Intergroup Emotions}},
  author={Garz{\'o}n-Velandia, Diana Camila and Barreto-Galeano, Mar{\'\i}a Idaly and Sabucedo-Cameselle, Jos{\'e} Manuel},
  journal={Analyses of Social Issues and Public Policy},
  volume={24},
  number={3},
  pages={621--644},
  year={2024},
  publisher={Wiley Online Library}
}

@article{kiesel2025affective,
    title = {{Affective Polarization in a Word: Open-Ended and Self-Coded Evaluations of Partisan Affect}},
    author = {Kiesel, Spencer and Amlani, Sharif},
    journal = {PLOS ONE},
    publisher = {Public Library of Science},
    year = {2025},
    month = {01},
    volume = {20},
    pages = {1-18},
    number = {1},

}

@article{suarez2022toxic,
  title={{Toxic Social Media: Affective Polarization After Feminist Protests}},
  author={Su{\'a}rez Estrada, Marcela and Juarez, Yulissa and Pi{\~n}a-Garc{\'\i}a, CA},
  journal={Social Media+ Society},
  volume={8},
  number={2},
  year={2022},
  publisher={SAGE Publications Sage UK: London, England}
}

@article{carpentras2023polarization,
  title={{How Polarization Extends to New Topics: An Agent-Based Model Derived from Experimental Data}},
  author={Carpentras, Dino and Lueders, Adrian and Maher, Paul J and O'Reilly, Caoimhe and Quayle, Michael},
  journal={Journal of Artificial Societies and Social Simulation},
  volume={26},
  number={3},
  year={2023},
  publisher={JASSS}
}

@inproceedings{jahan2024unraveling,
  title={{Unraveling the Tapestry of Deception and Personality: A Deep Dive into Multi-Issue Human-Agent Negotiation Dynamics}},
  author={Jahan, Nusrath and Mell, Johnathan},
  booktitle={Proceedings of the 23rd International Conference on Autonomous Agents and Multiagent Systems},
  pages={916--925},
  year={2024}
}

@article{brugiere2022handling,
  title={{Handling Multiple Levels in Agent-Based Models of Complex Socio-Environmental Systems: A Comprehensive Review}},
  author={Brugi{\`e}re, Arthur and Nguyen-Ngoc, Doanh and Drogoul, Alexis},
  journal={Frontiers in Applied Mathematics and Statistics},
  volume={8},
  pages={1020353},
  year={2022},
  publisher={Frontiers Media SA}
}

@article{liu2024more,
title={{A More Advanced Group Polarization Measurement Approach Based on LLM-Based Agents and Graphs}},
author={Liu, Zixin and Zhang, Ji and Ding, Yiran},
volume={30},
journal={Information Research an international electronic journal}, 
year={2025},
pages={93–107} }

@article{marchal2022nice,
  title={{“Be Nice or Leave Me Alone”: An Intergroup Perspective on Affective Polarization in Online Political Discussions}},
  author={Marchal, Nahema},
  journal={Communication Research},
  volume={49},
  number={3},
  pages={376--398},
  year={2022},
  publisher={Sage Publications Sage CA: Los Angeles, CA}
}

@article{ge2024scaling,
  title={{Scaling Synthetic Data Creation With 1,000,000,000 Personas}},
  author={Ge, Tao and Chan, Xin and Wang, Xiaoyang and Yu, Dian and Mi, Haitao and Yu, Dong},
  journal={arXiv preprint arXiv:2406.20094},
  year={2024}
}

@article{bail2018exposure,
  title={{Exposure to Opposing Views on Social Media Can Increase Political Polarization}},
  author={Bail, Christopher A and Argyle, Lisa P and Brown, Taylor W and Bumpus, John P and Chen, Haohan and Hunzaker, MB Fallin and Lee, Jaemin and Mann, Marcus and Merhout, Friedolin and Volfovsky, Alexander},
  journal={In Proceedings of the National Academy of Sciences},
  volume={115},
  number={37},
  pages={9216--9221},
  year={2018},
  publisher={National Academy of Sciences}
}

@article{santoro2022promise,
  title={{The Promise and Pitfalls of Cross-Partisan Conversations for Reducing Affective Polarization: Evidence from Randomized Experiments}},
  author={Santoro, Erik and Broockman, David E},
  journal={Science Advances},
  volume={8},
  number={25},
  pages={eabn5515},
  year={2022},
  publisher={American Association for the Advancement of Science}
}

@article{rossiter2023similar,
title = {{The Similar and Distinct Effects of Political and Non-Political Conversation on Affective Polarization}},
  author={Rossiter, Erin and Carlson, Taylor},
  journal={Preprint at https://erossiter. com},
  year={2023}
}

@article{iyengar2015fear,
  title={Fear and loathing across party lines: New evidence on group polarization},
  author={Iyengar, Shanto and Westwood, Sean J},
  journal={American journal of political science},
  volume={59},
  number={3},
  pages={690--707},
  year={2015},
  publisher={Wiley Online Library}
}

@article{jackson2017abm,
author = {Joshua Conrad Jackson and David Rand and Kevin Lewis and Michael I. Norton and Kurt Gray},
title ={Agent-Based Modeling: A Guide for Social Psychologists},
journal = {Social Psychological and Personality Science},
volume = {8},
number = {4},
pages = {387-395},
year = {2017},
}

@book{gilbert2008agent,
  title={Agent-Based Models},
  author={Gilbert, Nigel},
  volume={153},
  series= {Quantitative Applications in the Social Sciences},
  year={2008},
  publisher={SAGE Publications}
}

@article{Liu2024abmopinion,
    author = {Liu, Shan and Wen, Hao},
    title = {{Agent-Based Modelling of Polarized News and Opinion Dynamics in Social Networks: A Guidance-Oriented Approach}},
    journal = {Journal of Complex Networks},
    volume = {12},
    number = {4},
    year = {2024}
}

@article{gemini2024team,
  title={{Gemini: A Family of Highly Capable Multimodal Models}},
  author={Team, Gemini and Anil, Rohan and Borgeaud, Sebastian and Alayrac, Jean-Baptiste and Yu, Jiahui and Soricut, Radu and Schalkwyk, Johan and Dai, Andrew M and Hauth, Anja and Millican, Katie and others},
  journal={arXiv preprint arXiv:2312.11805},
  year={2024}
}

@article{unlu2024online,
  title={{Online Polarization and Identity Politics: An Analysis of Facebook Discourse on Muslim and LGBTQ+ Communities in Finland}},
  author={Unlu, Ali and Kotonen, Tommi},
  journal={Scandinavian political studies},
  volume={47},
  number={2},
  pages={199--231},
  year={2024},
  publisher={Wiley Online Library}
}

@article{harel2020normalization,
  title={{The Normalization of Hatred: Identity, Affective Polarization, and Dehumanization on Facebook in the Context of Intractable Political Conflict}},
  author={Harel, Tal Orian and Jameson, Jessica Katz and Maoz, Ifat},
  journal={Social Media+ Society},
  volume={6},
  number={2},
  pages={2056305120913983},
  year={2020},
  publisher={SAGE Publications Sage UK: London, England}
}

@article{arora2022polarization,
  title={{Polarization and Social Media: A Systematic Review and Research Agenda}},
  author={Arora, Swapan Deep and Singh, Guninder Pal and Chakraborty, Anirban and Maity, Moutusy},
  journal={Technological Forecasting and Social Change},
  volume={183},
  pages={121942},
  year={2022},
  publisher={Elsevier}
}

@article{de2024cross,
  title={{Cross-Partisan Discussions Reduced Political Polarization Between UK Voters, but Less So When They Disagreed}},
  author={de Jong, Jona F},
  journal={Communications psychology},
  volume={2},
  number={1},
  pages={5},
  year={2024},
  publisher={Nature Publishing Group UK London}
}

@article{xia2024integrated,
  title={{Integrated or Segregated? User Behavior Change after Cross-Party Interactions on Reddit}},
  author={Xia, Yan and Monti, Corrado and Keller, Barbara and Kivel{\"a}, Mikko},
  journal={arXiv preprint arXiv:2410.04923},
  year={2024}
}

@inproceedings{Harbar2025LLM,
author={Harbar, Yarolsav
and Chudziak, Jaros{\l}aw A.},
title={{Simulating Oxford-Style Debates with LLM-Based Multi-Agent Systems}},
booktitle={Intelligent Information and Database Systems},
year={2025},
publisher={Springer Nature Singapore},
address={Singapore},
pages={286--300},
}

@book{campbell1980american,
  title={{The American Voter}},
  author={Campbell, Angus},
  year={1980},
  publisher={University of Chicago Press}
}

@book{green2004partisan,
  title={{Partisan Hearts and Minds: Political Parties and the Social Identities of Voters}},
  author={Green, Donald P and Palmquist, Bradley and Schickler, Eric},
  year={2004},
  publisher={Yale University Press}
}

@article{martinez2024methodology,
  title={{Methodology for Measuring Individual Affective Polarization Using Sentiment Analysis in Social Networks}},
  author={Mart{\'\i}nez-Espa{\~n}a, Raquel and Fern{\'a}ndez-Pedauye, Julio and De Luc{\'\i}a, Jos{\'e} Giner-P{\'e}rez and Rojo-Mart{\'\i}nez, Jose Miguel and Bakdid-Albane, Kaoutar and Garc{\'\i}a-Escribano, Juan Jos{\'e}},
  journal={IEEE Access},
  year={2024},
  publisher={IEEE}
}

@inproceedings{saveski2022perspective,
  title={{Perspective-Taking to Reduce Affective Polarization on Social Media}},
  author={Saveski, Martin and Gillani, Nabeel and Yuan, Ann and Vijayaraghavan, Prashanth and Roy, Deb},
  booktitle={Proceedings of the International AAAI Conference on Web and Social Media},
  volume={16},
  pages={885--895},
  year={2022}
}

@article{yarchi2021political,
  title={{Political Polarization on the Digital Sphere: A Cross-Platform, Over-Time Analysis of Interactional, Positional, and Affective Polarization on Social Media}},
  author={Yarchi, Moran and Baden, Christian and Kligler-Vilenchik, Neta},
  journal={Political Communication},
  volume={38},
  number={1-2},
  pages={98--139},
  year={2021},
  publisher={Taylor \& Francis}
}

@article{lee2022social,
  title={{Social Media, Messaging Apps, and Affective Polarization in the United States and Japan}},
  author={Lee, Sangwon and Rojas, Hernando and Yamamoto, Masahiro},
  journal={Mass Communication and Society},
  volume={25},
  number={5},
  pages={673--697},
  year={2022},
  publisher={Taylor \& Francis}
}

@inproceedings{kamal2022,
author = {Kamal, Sadia and Gullic, Jade and Bagavathi, Arunkumar},
title = {{Modeling Polarization on Social Media Posts: A Heuristic Approach Using Media Bias}},
year = {2022},
publisher = {Springer-Verlag},
booktitle = {Proceeding of Foundations of Intelligent Systems: 26th International Symposium, ISMIS 2022},
pages = {35–43},
numpages = {9}
}

@article{wang2024decoding,
  title={Decoding echo chambers: Llm-powered simulations revealing polarization in social networks},
  author={Wang, Chenxi and Liu, Zongfang and Yang, Dequan and Chen, Xiuying},
  journal={arXiv preprint arXiv:2409.19338},
  year={2024}
}

@inproceedings{donkers2025understanding,
  title={Understanding Online Polarization Through Human-Agent Interaction in a Synthetic LLM-Based Social Network},
  author={Donkers, Tim and Ziegler, J{\"u}rgen},
  booktitle={Proceedings of the International AAAI Conference on Web and Social Media},
  volume={19},
  pages={457--478},
  year={2025}
}

@article{keijzer2024polarization,
  title={Polarization on social media: Micro-level evidence and macro-level implications},
  author={Keijzer, Marijn A and M{\"a}s, Michael and Flache, Andreas},
  journal={JASSS},
  volume={27},
  number={1},
  pages={7},
  year={2024},
  publisher={Particle Accelerator Society of Japan}
}

@article{jin2025synthetic,
  title={Synthetic Social Media Influence Experimentation Via an Agentic Reinforcement Learning Large Language Model Bot},
  author={Jin, Bailu and Guo, Weisi},
  journal={Journal of Artificial Societies and Social Simulation},
  volume={28},
  number={3},
  pages={1--6},
  year={2025},
  publisher={Journal of Artificial Societies and Social Simulation}
}

@inproceedings{harbar2025simulating,
  title={Simulating Oxford-Style Debates with LLM-Based Multi-Agent Systems},
  author={Harbar, Yarolsav and Chudziak, Jaros{\l}aw A},
  booktitle={Asian Conference on Intelligent Information and Database Systems},
  pages={286--300},
  year={2025},
  organization={Springer}
}

@inproceedings{gajewska2025leveraging,
  title={Leveraging a Multi-agent LLM-based system to educate teachers in hate incidents management},
  author={Gajewska, Ewelina and Wawer, Michal and Budzynska, Katarzyna and Chudziak, Jaroslaw A},
  booktitle={International Conference on Artificial Intelligence in Education},
  pages={332--339},
  year={2025},
  organization={Springer}
}

\end{document}